\documentclass[letterpaper, 10 pt, conference]{ieeeconf}  

\IEEEoverridecommandlockouts                              

\usepackage{graphicx}
\usepackage{booktabs}
\usepackage{hyperref}
\usepackage{amsmath}

\title{\LARGE \bf
Speech-Gesture GAN: Gesture Generation for Robots and Embodied Agents
}

\author{Carson Yu Liu$^{1}$, Gelareh Mohammadi$^{1}$, Yang Song$^{1}$ and Wafa Johal$^{2}$  
\thanks{$^{1}$Carson Yu Liu, Gelareh Mohammadi and Yang Song are with the Faculty of Engineering, School of Computer Science and Engineering,
        University of New South Wales, Sydney, NSW 2052, Australia
        {\tt\small carson.liu@.unsw.edu.au; g.mohammadi@unsw.edu.au; yang.song1@unsw.edu.au}}%
\thanks{$^{2}$Wafa Johal is with the Faculty of Engineering and Information Technology,
School of Computing \& Information Systems,
        University of Melbourne, Melbourne, VIC 3010, Australia
        {\tt\small wafa.johal@unimelb.edu.au}}%
}

\begin{document}

\maketitle

\begin{abstract}

Embodied agents, in the form of virtual agents or social robots, are rapidly becoming more widespread. In human-human interactions, humans use nonverbal behaviours to convey their attitudes, feelings, and intentions. Therefore, this capability is also required for embodied agents in order to enhance the quality and effectiveness of their interactions with humans. 
In this paper, we propose a novel framework that can generate sequences of joint angles from the speech text and speech audio utterances. Based on a conditional Generative Adversarial Network (GAN), our proposed neural network model learns the relationships between the co-speech gestures and both semantic and acoustic features from the speech input. In order to train our neural network model, we employ a public dataset containing co-speech gestures with corresponding speech audio utterances, which were captured from a single male native English speaker. 
The results from both objective and subjective evaluations demonstrate the efficacy of our gesture-generation framework for Robots and Embodied Agents.

\end{abstract}

\section{INTRODUCTION}

As a result of the ongoing improvement of humanoid robots and computer graphics, conversational embodied agents, including social robots and virtual agents, have emerged as effective interaction instrumentality. The ESI (Evaluation of Social Interaction) \cite{fisher2010evaluation}, a human evaluation instrument, identifies important social skills such as approaching, speaking, turn-taking, gazing and gesturing. Therefore, in human-agent interactions, social agents also need these social capabilities similar to humans. In particular, human gestures are a form of nonverbal cues utilised with utterances in interpersonal interaction. Secondly, researchers revealed that in certain cultures, speech and gestures are tightly linked in time \cite{streeck1993gesture}. Therefore, it is crucial to create gestures and briefly integrate them with speech while designing embodied agents. In fact, the danger for embodied agents is a mismatch between verbal and nonverbal information, which may cause extraordinary unpleasantness to the communicators \cite{johal2015non}. Thirdly, gestures may be used to emphasise words, demonstrate purpose, depict things more vividly, and aid understanding of a conversation \cite{dick2009co}. In human-robot interaction, it has been discovered that common language gestures strengthen the robot's attraction and prospective contact motivation \cite{hamacher2016believing}. However, considering the diversity of embodied agents and the physical limits of robots, it does not seem feasible to manually create gestures for each possible speech.  Linguists \cite{mcneill1992hand} suggested a categorisation system with four classes: 1) Iconic (expressing an object's features or behaviours); 2) Deictic, or pointing (indicating an object's position); 3) Metaphoric (representing abstract concepts with a concrete form); 4) Beat (keeping with the rhythm of speech). Only the Beat gestures are audio signal-dependent (speech acoustic), while other types of gestures rely on the speech context (speech semantic). Therefore, gesture generation frameworks with single modal input can lack some types of gestures.

Motivated by the accomplishments of GAN (Generative Adversarial Network) \cite{goodfellow2014generative} in generative models, we propose a GAN-structured neural network model to generate gestures from speech. We trained our model on a gesture dataset with English speech. The subjective evaluation demonstrates the proposed model is effective, showing a good performance when compared with the ground truth. Also, the objective evaluation results confirm our model is highly effective when compared with other state-of-the-art gesture generation models.

The contribution of our work is two-fold: 1) We propose a novel GAN-based generative framework that can use multimodal inputs to extract semantic and acoustic features as conditional information for adversarial training and generate multiple gestures from the same speech input using different input noises. 2) A comprehensive evaluation of the full model from objective to subjective with ablation studies of the outcomes of various designs and crucial modelling options;

The rest of this paper is organised as follows: We first introduce the background and related work in Section II. Then, Section III describes our proposed speech-based gesture generation framework, including features extraction, model architecture and its implementation. Next, sections IV and V explain the quantitative metrics used in our proposed model and quantitative result with validation on an extra user study. Finally, we conclude our work with a brief discussion.

\section{Related Work}
Several gesture generation approaches, ranging from rule-based to innovative data-driven, have been created in recent years. Initially, most approaches were rule-based; however, rule-based approaches result in a repetitious and monotonous experience in the lifetime human-agent connection. Recent innovative approaches are data-driven, enabling more variety in gesture production but making it more difficult to adapt to the physical limits of the embodied agents.

\subsection{Rule-based gesture generation}
The primary concept behind rule-based generation approaches is to correlate speech syllables, and words with gestures as a straightforward way to produce gestures from speech content \cite{bremner2009beat}. The rules for generating gestures in these studies were hand-defined by specialists \cite{kim2012automated,kim2012gesture,mlakar2013tts}. One study \cite{kim2012automated} derived punctuation marks from a sentence using a dialogue sentence analysis methodology. Using image processing and clustering approaches, unique research \cite{kadono2016generating} produced its own gestures dictionary for speech gestures from internet images, although the processing of gesture production is still governed by rules. For rule-based gesture generation, the greatest drawback is that manually defining a gesture pattern for each word requires an enormous amount of time and effort. By utilising the machine learning technique, the issue of repeated and labor-intensive generation of a speech gestures dictionary might be addressed.

\subsection{Data-driven gesture generation}
Recent Data-driven studies focus on learning mapping functions from speech text or speech audio or both of them to speech gestures.

\subsubsection{Gesture Generation with Speech Text}
Yoon et al. \cite{yoon2019robots} presented a seq2seq-based autoencoder model which employed speech text as input to generate 2D co-gestures; they also implemented their model on the NAO robot. Another work \cite{ishi2018speech} also extracted speech text features as input for their probabilistic model. However, both of them observed an unusual mapping issue in which the synthesised audio and produced gestures could not be closely synchronised. 

\subsubsection{Gesture Generation with Speech Audio}
Hasegawa et al. \cite{hasegawa2018evaluation} extracted the MFCCs (Mel- Frequency Cepstral Coefficients) from the inputted audio as the speech representation; they used a bi-directional LSTM (Long Short-Term Memory) based recurrent neural network to generate co-speech gestures and then went through a noise filter as smoothing step. With the same speech gesture database, Kucherenko et al. \cite{kucherenko2019analyzing} presented an autoencoder, which is used for representation learning to align the audio with gestures. Ferstl et al. \cite{ferstl2019multi} also used bi-directional LSTM regression with adversarial training to generate gestures from acoustic features (MFCCs with audio pitch); they also utilized multiple discriminators in adversarial training to improve the results from the generator. Our proposed approach varies from prior systems in that it generates co-speech gestures using text transcription and audio utterances.

\subsubsection{Gesture Generation with Multimodal Input}
Single modality systems have clear limitations; as mentioned before, the lack of either acoustic or semantic features resulting from the single modal input is currently a significant hurdle to achieving outstanding results. However, multimodality systems could address this problem. Kucherenko et al. \cite{kucherenko2020gesticulator} proposed the first multimodal input autoregressive neural network model on co-speech gesture generation. Yoon et al. \cite{yoon2020speech} added speaker identity as third modal input to achieve style control.

\section{Proposed Speech-based Gesture Generation Framework}

\subsection{Speech and Gesture Dataset}
Unlike previous studies that used non-English gesture datasets \cite{takeuchi2017speech}, small gesture datasets \cite{chiu2011train}, datasets with low-quality gestures \cite{yoon2019robots} or multi-language datasets \cite{Liu2022} our proposed speech-based gesture generation framework is specifically trained with the Trinity Dataset \cite{ferstl2018investigating}, that captured from a single male actor who is an English native speaker with 20 Vicon cameras (a sort of motion capture cameras). This gesture dataset contains 244 minutes of speech and gesture data from among a variety of topics, e.g., daily activities, hobbies and movies. First, we removed lower body data, because our work is aiming at co-speech gestures. Then, in order to save training time, for the upper body data, we used 4 joints from the spine, 2 joints from the neck, 3 joints from the left and right side arm, 2 joints from both side shoulder, 1 joint from the head. In addition, the fingers data is removed due to two reasons: 1) poor quality of data and 2) many common humanoid robots like NAO and Pepper from Softbank Robotics do not have enough fingers like human beings. Finally, we have speech audio utterances in the form of 44 kHz Waveform Audio file format, and speech text transcripts in the form of JavaScript Object Notation file format with timestamps and corresponding gestures in the form of Biovision Hierarchy file format.

\subsection{Data Pre-processing and Feature Extraction}
Based on the experiments of previous work\cite{kucherenko2020gesticulator}, we employ frame synchronization at 20 FPS (Frames Per Second) during feature extraction.
The gesture data in Biovision Hierarchy consist of Euler angles and offsets of each joint in a hierarchical structure. Unlike previous studies that adopted conversion of Euler angles and absolute position in 3D coordinates \cite{wu2021modeling}, we converted Euler angles to exponential maps \cite{grassia1998practical} because it is easy to convert exponential maps back to Euler angles and will not introduce potential discontinuities issues. After frame conversion from 60 FPS to 20 FPS, we get 45 features for each frame of gestures.

As for the acoustic features extraction, similar to other state-of-the-art in speech-based gesture generation\cite{hasegawa2018evaluation,kucherenko2019analyzing,ahuja2020style,alexanderson2020style}, in order to align with the gesture features, we get feature vectors in 26 dimensions (for 26 Mel-spaced filterbanks) by calculating the MFCCs of the audio utterances waveform with the same frame rate, which is a representation of an utterance's short term power spectrum.

However, a sequence of the speech audio utterances and its corresponding speech text transcripts generally have different lengths. In order to address this problem, we first encoded the words with semantic information as 768-dimensional vectors by using the BERT\cite{devlin2018bert} pre-trained model, a state-of-the-art neural network model that uses surrounding text to assist computers in grasping the meaning of ambiguous words in a text. As for the words that do not have semantic information, we encoded them as fixed vectors that have the same dimensions as the BERT features. Then, we used the exact utterance time information of each world to upsample the text features. Therefore, the text and audio feature sequences get aligned and uniform.

\subsection{Problem Formulation}
The problem of the co-gesture generation from speech can be defined as a mapping function $\textbf{F}_{Generation}$, which is shown in Equation \ref{equ:problem} for a segment of the input speech length $T$, where $\textbf{s}_{a}=[s_{a}]_{t=1:T}$ are the features extracted from speech audio utterances. Likewise, the features extracted from speech text are $\textbf{s}_{t}=[s_{t}]_{t=1:T}$, with multiple noise $\textbf{n}$. The corresponding result $\textbf{g}=[\text{g}_{t=1:T}]$ can be a sequence of Euler angles of selected joints in the form of $\text{g}_{t}=[pitch_{t}^{i}, raw_{t}^{i}, yaw_{t}^{i}]_{i=1:J}$, where $J$ is the number of selected joints. Furthermore, we define $\textbf{g}=[\text{g}_{t=1:T}]$ as a sequence of 3D (three-dimensional) coordinates of selected joints, with $\text{g}_{t}=[x_{t}^{i}, y_{t}^{i}, z_{t}^{i}]_{i=1:J}$. The object of our problem is to achieve the maximization of the conditional probability $p(\textbf{g}|\textbf{s})$ to match well with the given speech input, where $\textbf{s}$ is the concatenation of $\textbf{s}_{a}$ and $\textbf{s}_{t}$. 

\begin{equation}
    \textbf{g}=\textbf{F}_{Generation}(\textbf{s}_{a},\textbf{s}_{t},\textbf{n})
    \label{equ:problem}
\end{equation}

\subsection{Model Architecture}
\begin{figure}[!htbp]
    \centering
    \includegraphics[scale=0.56]{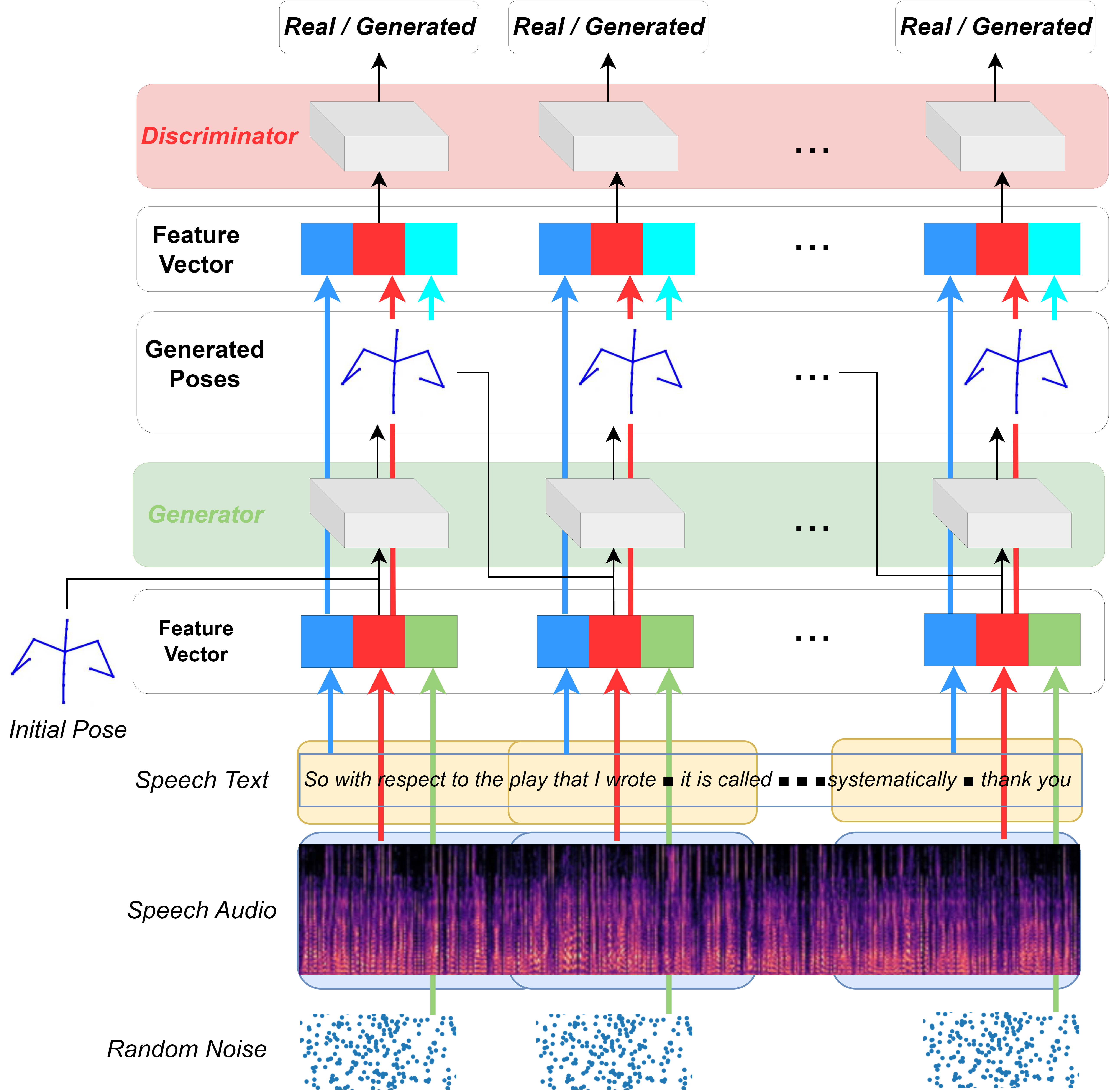}
    \caption{The architecture of the proposed gesture generation model}
    \label{fig:model}
\end{figure}

Speech features extracted from audio utterances and text transcripts are used as the condition in our proposed model, which is a conditional GAN-based architecture. Figure \ref{fig:model} shows the overview of the architecture.

In the generation step, a random noise $\textbf{n}$ from a normal distribution is reproduced in the same length as the speech features. Then, the noise, the text embeddings $\textbf{s}_{t}$  and the MFCCs values $\textbf{s}_{a}$ are concatenated as the feature vector, take it into the generator to get the corresponding sequence of gestures. Specifically, we employed the initial pose for the previous frames to improve the continuity during gesture generation. In order to improve the generator, we concurrently trained the discriminator to calculate the difference between the real distribution and fake distribution on the speech features condition. Next, after getting the sequence of generated gestures, we concatenated the generated gestures or real gestures with accompanying audio and semantic features and then sent them into the discriminator. The output value shows if the input gestures were real or fake for the corresponding speech features condition.

\subsection{Gesture Generator}
\begin{figure}[!htbp]
    \centering
    \includegraphics[scale=0.65]{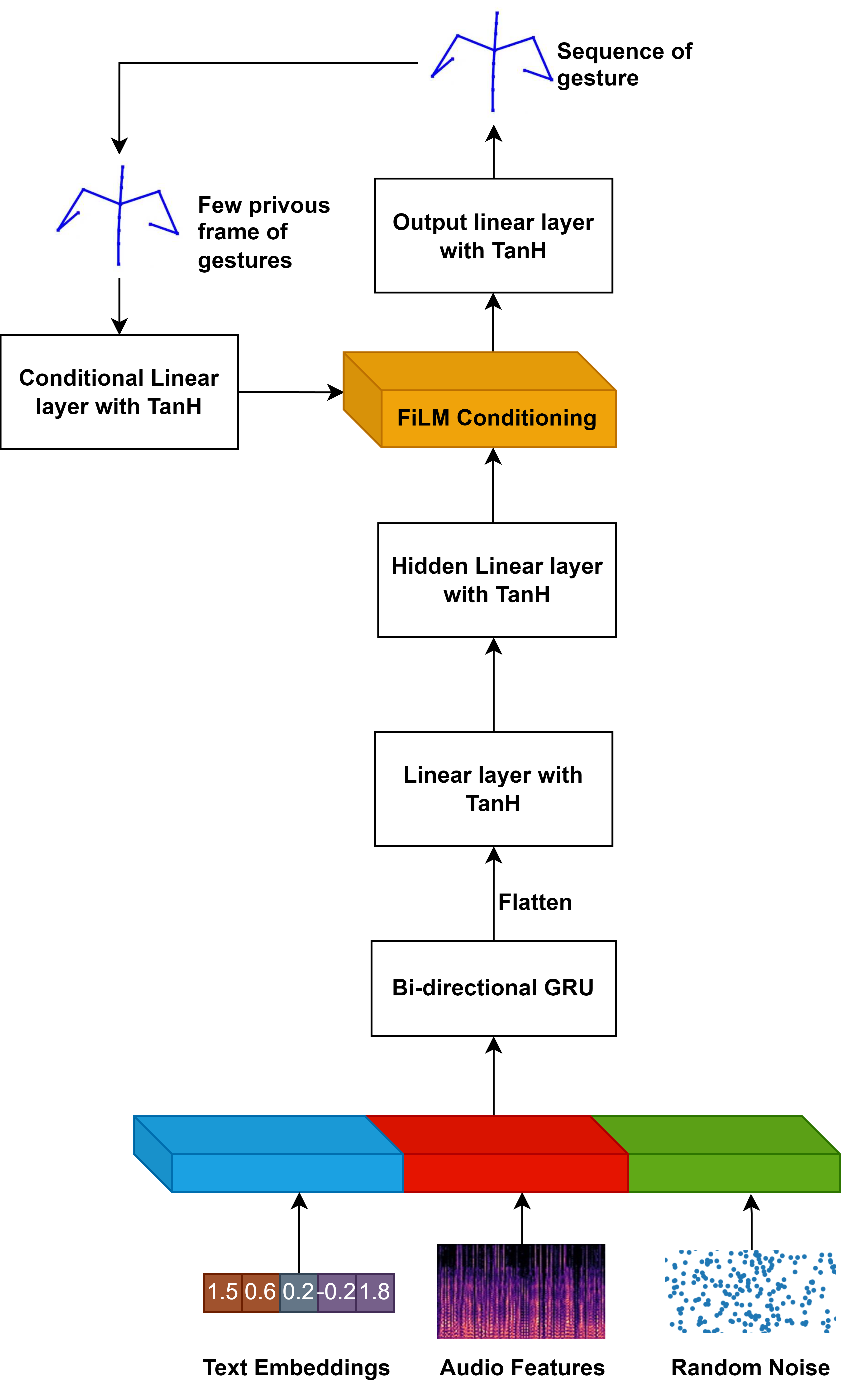}
    \caption{Gesture Generator}
    \label{fig:generator}
\end{figure}

Our gesture generator $G$ generates gestures using encoded semantic and acoustic features as input. The structure of the generator $G$ is shown in Figure \ref{fig:generator}. 

\begin{table}[h]
\caption{Generator architecture}
\label{tab:Generator architecture}
\begin{center}
\begin{tabular}{ c | l  l}
\toprule
\specialrule{0em}{1pt}{1pt}
\textbf{Layer} & \textbf{Layer Type} & \textbf{Hyperparameter} \\
\midrule
\specialrule{0em}{1pt}{1pt}
1 &  GRU &  $C_{in}=814 , C_{out}=128, L_{num}=2 $\\
\specialrule{0em}{1pt}{1pt}
2 & Linear & $C_{in}=3840 , C_{out}=512 $\\
\specialrule{0em}{1pt}{1pt}
3 & Linear & $C_{in}=135 , C_{out}=512 $\\
\specialrule{0em}{1pt}{1pt}
4 & Linear & $C_{in}=512 , C_{out}=256$\\
\specialrule{0em}{1pt}{1pt}
5 & Linear & $C_{in}=256 , C_{out}=45 $\\
\specialrule{0em}{1pt}{1pt}
\bottomrule
\end{tabular}
\end{center}
\end{table}

First, we concatenate the text embedding, MFCCs and random noise as a long vector, then send them through to the two-layer bi-direction GRU (Gated recurrent unit) with 0.2 dropouts. Next, the vector passes through the following linear layer with the TanH activation function to reduce the dimensionality of the feature. In order to ensure the continuity of generated gestures, we used the few frames of previously generated gestures as condition information to feed back to the FiLM (Feature-wise Linear Modulation) layer \cite{perez2018film}, as another state-of-the-art work \cite{kucherenko2020gesticulator} did. Finally, the output layer is a linear layer with the TanH activation function to get a possible range of results. The layers detail of the gesture generator is shown in Table \ref{tab:Generator architecture}, where $C_{in}$, $C_{out}$ are dimensions of in and out channels, and $L_{num}$ is the number of GRU layers.

\subsection{Adversarial Scheme}
\begin{figure}[!htbp]
    \centering
    \includegraphics[scale=0.65]{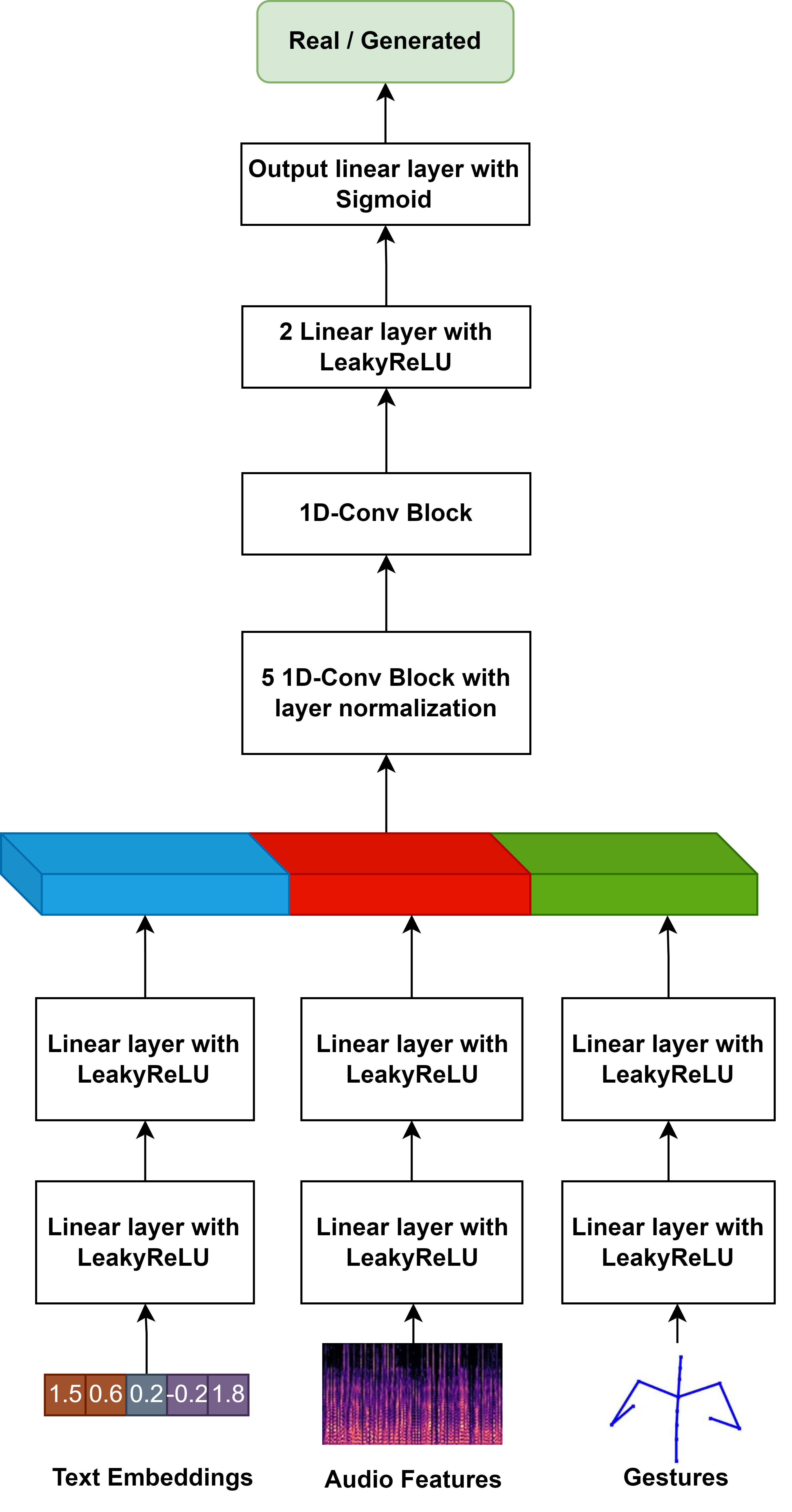}
    \caption{Discriminator}
    \label{fig:discriminator}
\end{figure}

In order to optimize our gesture generator, a discriminator $D$ is used in our adversarial scheme. Figure \ref{fig:discriminator} illustrates the structure of our discriminator. First, the sequence of generated gestures from the generator, text embeddings and MFCCs both individually go through two linear layers: one with the Leaky ReLU activation function and the next one without the activation function. Inspired by the work \cite{vougioukas2018end}, we take the vector of the concatenated gestures, audio and text features and then feed them into five layers of the 1D convolutional block, which consists of one 1D convolutional layer with Leaky ReLU and layer normalization, finally followed by an extra 1D convolutional layer. Next, the vector passes through two linear layers with Leaky ReLU for vector dimensional reduction. At the end of the discriminator, using a sigmoid activation function, the result is compressed between 0 and 1. These values could determine if the input gestures are real and well-matched with the condition features. The layers detail of the discriminator is shown in Table \ref{tab:Discriminator architecture}, where $k$, $s$ and $p$ are kernel size, stride and padding, respectively.

\begin{table}[h]
\caption{Discriminator architecture}
\label{tab:Discriminator architecture}
\begin{center}
\begin{tabular}{ c | l  p{5.5cm}}
\toprule
\specialrule{0em}{1pt}{1pt}
\textbf{Layer} & \textbf{Layer Type} & \textbf{Hyperparameter} \\
\specialrule{0em}{1pt}{1pt}
\midrule
\specialrule{0em}{1pt}{1pt}
1 & Linear & $C_{in}=768 , C_{out}=32 $\\
\specialrule{0em}{1pt}{1pt}
1 & Linear & $C_{in}=26 , C_{out}=32 $\\
\specialrule{0em}{1pt}{1pt}
1 & Linear & $C_{in}=45 , C_{out}=32 $\\
\specialrule{0em}{1pt}{1pt}
2 & Linear & $C_{in}=32 , C_{out}=64 $\\
\specialrule{0em}{1pt}{1pt}
3 & 1D-Conv & $k=3,s=1,p=0,C_{in}=192,C_{out}=192 $\\
\specialrule{0em}{1pt}{1pt}
4 & 1D-Conv & $k=4,s=2,p=0,C_{in}=192,C_{out}=256 $\\
\specialrule{0em}{1pt}{1pt}
5 & 1D-Conv & $k=3,s=1,p=0,C_{in}=256,C_{out}=256 $\\
\specialrule{0em}{1pt}{1pt}
6 & 1D-Conv & $k=4,s=2,p=0,C_{in}=256,C_{out}=512 $\\
\specialrule{0em}{1pt}{1pt}
7 & 1D-Conv & $k=3,s=1,p=0,C_{in}=512,C_{out}=512 $\\
\specialrule{0em}{1pt}{1pt}
8 & 1D-Conv & $k=4,s=2,p=0,C_{in}=512,C_{out}=1024 $\\
\specialrule{0em}{1pt}{1pt}
9 & Linear & $C_{in}=1024 , C_{out}=512 $\\
\specialrule{0em}{1pt}{1pt}
10 & Linear & $C_{in}=512 , C_{out}=256 $\\
\specialrule{0em}{1pt}{1pt}
11 & Linear & $C_{in}=256 , C_{out}=1 $\\
\specialrule{0em}{1pt}{1pt}
\bottomrule
\end{tabular}
\end{center}
\end{table}

\subsection{Training}
The losses listed below are used to train the proposed framework. The gesture generator is trained by using the loss $\textbf{L}_{G}$ in Equation \ref{equ:lg}, while the loss $\textbf{L}_{D}$ in Equation \ref{equ:ld} is used for training the discriminator.

\begin{equation}
    \textbf{L}_{G} = \alpha \cdot \textbf{L}_{G}^{mse} + \beta \cdot \textbf{L}_{G}^{continuity} + \lambda \cdot \textbf{L}_{G}^{WGAN}
    \label{equ:lg}
\end{equation}

\begin{equation}
    \textbf{L}_{G}^{mse} = \frac{1}{n}\sum_{i=1}^{n}(\textbf{g}_{i}-\hat{\textbf{g}}_{i})^{2}
    \label{equ:lg_mse}
\end{equation}

\begin{equation}
    \textbf{L}_{G}^{continuity} = \frac{1}{n}\sum_{i=1}^{n}(\textbf{S}_{i}-\hat{\textbf{S}}_{i})^{2}
    \label{equ:lg_con}
\end{equation}

\begin{equation}
    \textbf{L}_{G}^{WGAN} = -\frac{1}{N}\sum_{i=1}^{n}D(\textbf{s}_a,\textbf{s}_t,\hat{\textbf{g}}_{i})
    \label{equ:lg_wgan}
\end{equation}

\begin{equation}
    \textbf{L}_{D}=\frac{1}{N}\sum_{i=1}^{n}D(\textbf{s}_a,\textbf{s}_t,\hat{\textbf{g}}_{i})-\frac{1}{N}\sum_{i=1}^{n}D(\textbf{s}_a,\textbf{s}_t,\textbf{g}_{i})
    \label{equ:ld}
\end{equation}

Where $\textbf{s}_a$, $\textbf{s}_t$ represent the speech audio and text features, respectively. Specifically, $n$ is the total duration of the gesture sequence, $\textbf{g}_{i}$ and $\hat{\textbf{g}}_{i}$ are the $i$th original gesture and $i$th generated gesture, respectively. Using MSE (mean squared error) in Equation \ref{equ:lg_mse} and continuity loss in Equation \ref{equ:lg_con}, we reduced the gap between original gestures in training samples and the matching generated gestures while training our gesture generator. This loss $\textbf{L}_{G}^{continuity}$ can be construed as the mean squared error for the current speed difference of $i$th original gesture speed $\textbf{S}_{i}$ and $i$th generated gesture speed $\hat{\textbf{S}}_{i}$. The adversarial losses $\textbf{L}_{G}^{WGAN}$ in the Equation \ref{equ:lg_wgan}, where $G$ is the generator and $\textbf{L}_{D}$ where $D$ is discriminator come from the WGAN (Wasserstein Generative Adversarial Networks) \cite{arjovsky2017wasserstein}, an improved generative model that makes the training more stable when compared with the traditional GAN model. As in the GAN training, $\textbf{L}_{G}$ and $\textbf{L}_{D}$ are alternately used to update the gesture generator and discriminator. The trained result of $D()$ is 1 for original gestures and 0 for generated (fake) gestures.

We split the Trinity dataset into three parts: 84\% for the training set (205 minutes), 7.4\% for the validation set (18 minutes), and 8.6\% for the test set (21 minutes), and every set has its own audio, text transcript, and co-speech motion files.
We trained the proposed model for 100 epochs. The batch size was 64, while the learning rate was 0.0001. The optimizer for both gesture generator and the discriminator is Adam with $\beta_{1}=0.9$ and $\beta_{2}=0.999$. The weights for loss functions ($\alpha=1$, $\beta=0.6$, $\lambda=0.3$) were set experimentally. The model was trained for approximately 7 hours on a GPU (NVIDIA RTX 3070) with CPU (Intel 12900k). For a 30-second speech input, the overall compilation time from loading the speech input to feature extraction to final motion file generation takes about 12.3 seconds in total, either by loading the pre-trained model on CPU or loading it on GPU.

\section{Evaluation Metrics}
\subsection{Quantitative Evaluation}

Objectively evaluating generated gestures is difficult due to the lack of suitable measures for measuring the perceived quality of co-speech gestures. The recent review works \cite{liu2021speech, wolfert2022review} indicated that there is no consensus in previous works on which quantitative evaluation could be applied to assess the quality of the generated gestures. Objective assessment measures are still necessary for fair and trustworthy comparisons of various models. We mainly utilised measurements suggested by earlier studies as a trend towards developing standard assessment measures in the area of gesture generation. Thus, we followed the step from the state-of-the-art model Gesticulator \cite{kucherenko2020gesticulator}, used Acceleration of gestures, Jerk of gestures which is Acceleration changing rate to evaluate the average value of a sequence of gestures. Additionally, RMSE (Root-Mean-Square Error) is also included, which is a common metric of the discrepancies between results produced by a model, shown in the equation \ref{equ:rmse}:

\begin{equation}
    \textbf{RMSE}(\textbf{g}_{i},\hat{\textbf{g}}_{i})=\sqrt{\frac{1}{n}\sum_{i=1}^{n}(\textbf{g}_{i}-\hat{\textbf{g}}_{i})^{2}}
    \label{equ:rmse}
\end{equation}

Where $\textbf{g}_{i}$ and $\hat{\textbf{g}}_{i}$ are the coordinates of $i$th original gestures and $i$th generated gesture, respectively. The $n$ is the total length of the sequence of a gesture. 

\subsection{User Study}
The most important purpose in the field of gesture generation is to produce gestures that make humans feel comfortable and natural in communication. However, the quantitative metrics are difficult to evaluate these characteristics, which need human perception. For example, some gestures that score very low in objective evaluations may look natural and comfortable. Hence, we conducted a user study to measure the generated gestures against the ground truth.

Three criteria are used in our user study, including naturalness, time consistency and semantics of the gestures.  As shown in Table \ref{tab:user study}, we used the three questions for each criterion, which is frequently used in other works \cite{hasegawa2018evaluation, kucherenko2019analyzing, kucherenko2020gesticulator}.

\begin{table}[!htbp]
\caption{The criterion used in user study}
\label{tab:user study}
\begin{center}
\begin{tabular}{ l | l }
\toprule
\specialrule{0em}{1pt}{1pt}
\textbf{Criterion} & \textbf{Description}\\
\specialrule{0em}{1pt}{1pt}
\midrule
\specialrule{0em}{1pt}{1pt}
 & Gestures were natural\\ 
\specialrule{0em}{1pt}{1pt}
Naturalness & Gestures were smooth\\ 
\specialrule{0em}{1pt}{1pt}
 & Gestures were comfortable \\ 
\specialrule{0em}{1pt}{1pt}
\hline
\specialrule{0em}{1pt}{1pt}
 & Gestures timing was matched to speech\\ 
 \specialrule{0em}{1pt}{1pt}
Time Consistency & Gesture speed was matched to speech\\ 
\specialrule{0em}{1pt}{1pt}
 & Gesture pace was matched to speech \\ 
 \specialrule{0em}{1pt}{1pt}
\hline
\specialrule{0em}{1pt}{1pt}
 & Gestures were matched to speech content\\ 
 \specialrule{0em}{1pt}{1pt}
Semantics &  Gesture well described speech content\\ 
\specialrule{0em}{1pt}{1pt}
 & Gesture helped people to understand the content \\
 \specialrule{0em}{1pt}{1pt}
\bottomrule
\end{tabular}
\end{center}
\end{table}

\section{Results}

\subsection{Objective Evaluation}

\begin{figure*}[!htbp]
    \centering
    \includegraphics[scale=0.5]{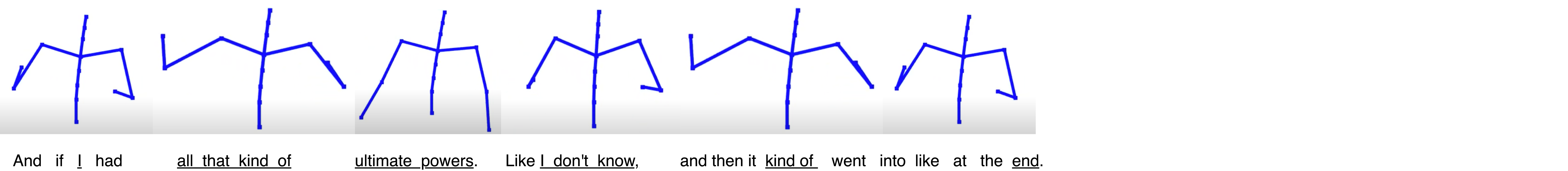}
    \caption{Qualitative results.}
    \label{fig:motion}
\end{figure*}

In order to benchmark against the state-of-the-art model, we compared our proposed model with the Gesticulator \cite{kucherenko2020gesticulator}, the first multimodality speech gesture generation model. As shown in Table \ref{tab:objective}, the results are averaged values of Acceleration, Jerk and RMSE over 50 samples from the original test dataset. Fig \ref{fig:motion} shows an example. In accordance with the circumstances of the speech, various gestures are used. It depicts a metaphorical motion of spreading the arms to represent the idea of "all that kind of" and "kind of". The iconic gestures that depict "ultimate powers" is generated by the framework. For the speech "don't know", the skeleton makes the shrug to depict an iconic gesture. The framework correctly recognised characteristic words and produced a deictic gesture for "I" and "end".

\begin{table}[!htbp]
\caption{Objective evaluation of proposed model with the state-of-the-art. For Metrics: Closer to the Ground Truth is better. Acceleration(Acc.).}
\label{tab:objective}
\begin{center}
\begin{tabular}{p{1.9cm} p{1.6cm} p{2.1cm} p{1.3cm}}
\toprule
\specialrule{0em}{1pt}{1pt}
\textbf{Model} & \textbf{Acc.}$(cm/s^2)$ & \textbf{Jerk}$(cm/s^3)$ & \textbf{RMSE}$(cm)$ \\
\specialrule{0em}{1pt}{1pt}
\midrule
\specialrule{0em}{1pt}{1pt}
Gesticulator & $63.8 \pm 8.3$ & $1332 \pm 192$ & $13.0 \pm 14.7$\\
\specialrule{0em}{1pt}{1pt}
Proposed Model & $94.48 \pm 19.64$ & $2187.76 \pm 611.97$ & $4.21 \pm 4.54$\\
\specialrule{0em}{1pt}{1pt}
\hline
\specialrule{0em}{1pt}{1pt}
Ground Truth & $144.7 \pm 36.6$ & $2322 \pm 538$ & $0$\\
\specialrule{0em}{1pt}{1pt}
\bottomrule
\end{tabular}
\end{center}
\end{table}

\subsection{Subjective Evaluation}

Our user study was delivered via an anonymous online questionnaire with video clips\footnote{\href{https://www.youtube.com/watch?v=6N3--ARpI4Q}{Sample from proposed group} and \href{https://www.youtube.com/watch?v=Dn9eE7y68sc}{sample from GT group}}. The questionnaire asked participants to rate the statements from strongly disagree value (1) to strongly agree value 
(7) after watching gesture videos. We made 10 sets of videos by using different speeches. Each set contains two 10s video clips: ground truth and generated gestures from our proposed model. The order in which the videos appear is random, and the entire questionnaire takes about 15 minutes to complete. Our user study is supported by UNSW Research Ethics Compliance Support \footnote{HC No: HC220411}. From the social media, 20 native English speakers (13 male, 7 female, mean = 24.1, standard deviation = 1.8 years old) participated in our user study. Fig \ref{fig:user study} below presented the results.

\begin{figure}[!htbp]
    \centering
    \includegraphics[scale=0.47]{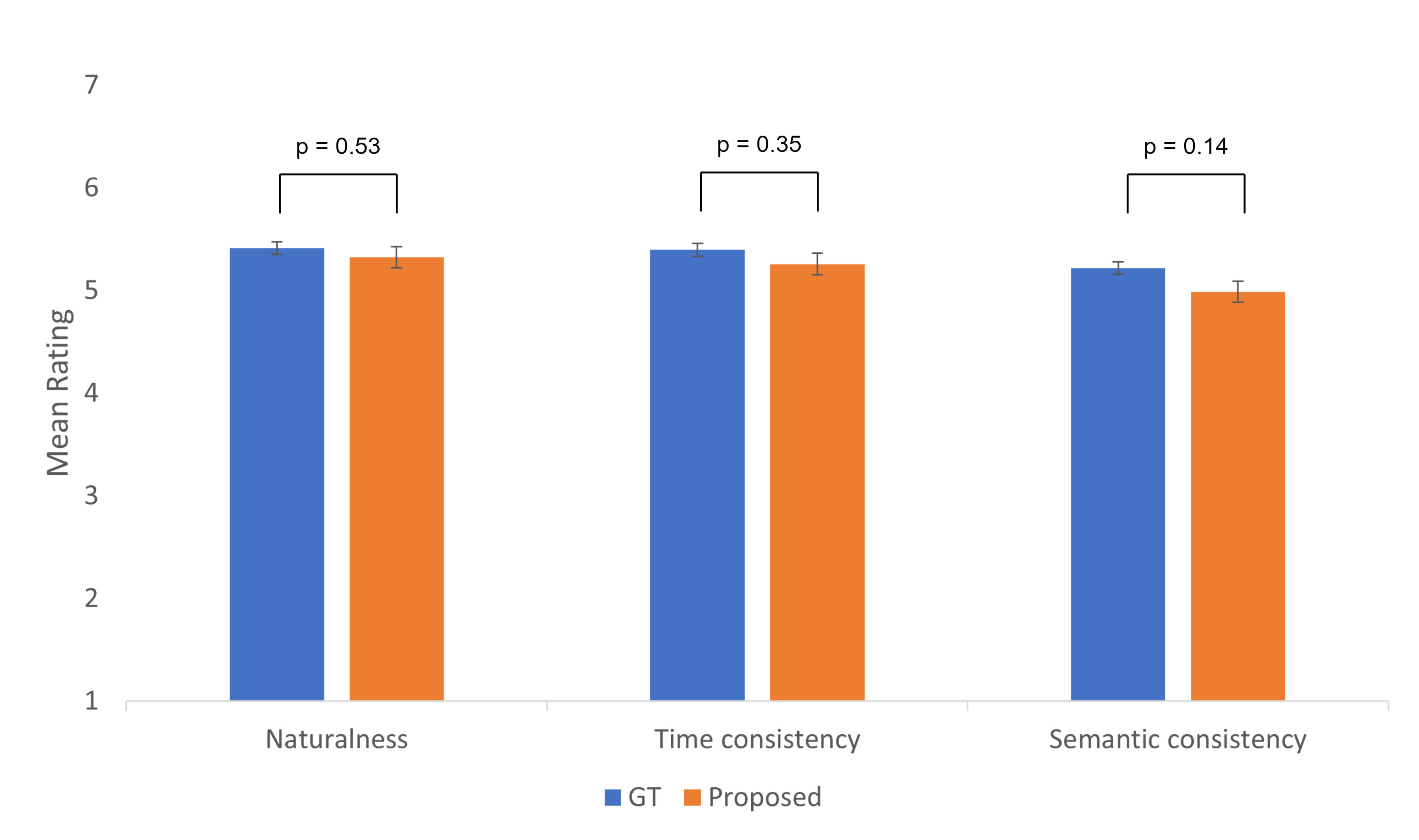}
    \caption{Results of the user study}
    \label{fig:user study}
\end{figure}

A two-tailed T-test was used to determine if there was a statistically significant difference in the scores of the GT and proposed groups. Although the mean rating scores of the proposed model are lower than the ground truth, especially in semantic consistency, there was no statistically significant difference among these three criteria. For the naturalness, between the ground truth group (M = 5.41, SD = 1.52) and the proposed group (M = 5.33, SD = 1.56), 
$t$ = 0.6210, $p$ = 0.5349, and the result is not significant at $p$ $<$ 0.05. For the time consistency, between the ground truth group (M = 5.40, SD = 1.64) and the proposed group (M = 5.26, SD = 1.59), 
$t$ = 0.9317, $p$ = 0.3520, and the result is not significant at $p$ $<$ 0.05. For the semantic consistency, between the ground truth group (M = 5.22, SD = 1.73) and the proposed group (M = 4.99, SD = 1.70), 
$t$ = 1.48494, $p$ = 0.1382, and the result is not significant at $p$ $<$ 0.05. 

Overall, by conventional criteria, we select a significance threshold of $p$ value: 0.05. We observed all $p$ values of different criteria are more than 0.05. Then, it indicated the difference between the means of the proposed model and the ground truth is not probably the result of chance. We have no basis in the data to infer that the population means of the proposed model and GT group are different because of the lack of proof of difference. Hence, the difference is considered to be not statistically significant. The results mean the performance of the proposed model is similar to the ground truth.

\section{Ablation Study}
In this section, we conducted two ablation studies. One is to evaluate the difference between various input speech features, and another is to focus on the various framework structures. Both of them are evaluated by objectively.

\subsection{Audio Features Experiments}
According to the previous data-driven method for gesture generation, they tend to use MFCCs, prosodic and Mel spectrogram as speech audio features.
In order to get a better understanding of the impact of the audio feature's type, we proposed five models that used different features input. Detail settings and results are shown in Table \ref{tab:audio ablation}.  

\begin{table}[!htbp]
\caption{Objective evaluation of audio features. For Metrics: Closer to the Ground Truth is better. Acceleration(Acc.).}
\label{tab:audio ablation}
\begin{center}
\begin{tabular}{p{1.96cm} p{1.7cm} p{2.1cm} p{1.3cm}}
\toprule
\textbf{Model} & \textbf{Acc.}$(cm/s^2)$ & \textbf{Jerk}$(cm/s^3)$ & \textbf{RMSE}$(cm)$ \\
\midrule
MFCCs & $94.48 \pm 19.64$ & $2187.76 \pm 611.97$ & $4.21 \pm 4.54$\\
Mel Spectrogram & $69.18 \pm 12.90$ & $1234.70 \pm 234.05$ & $4.50 \pm 5.12$\\
Prosodic & $60.54 \pm 8.90$ & $948.00 \pm 138.98$ & $4.42 \pm 4.93$\\
MFCCs + Prosodic & $94.99 \pm 22.88$ & $2157.39 \pm 607.94$ & $4.28 \pm 4.69$\\
Mel Spectrogram + Prosodic & $70.39 \pm 13.45$ & $1273.65 \pm 242.76$ & $4.29 \pm 4.77$\\
\hline
Ground Truth & $144.7 \pm 36.6$ & $2322 \pm 538$ & $0$\\
\bottomrule
\end{tabular}
\end{center}
\end{table}

Same as the quantitative measurement, we trained 100 epochs for each type of model. From the results, we found the MFCCs-based model got the best result in RMSE and Jerk metrics, and a suboptimal result in the Acceleration metric. Although the MFCCs + Prosodic-based model achieved the best performance in the Acceleration metric when compared with the ground truth, it only showed slightly higher than the MFCCs-based model. Hence, MFCCs based model is the best one, which is much closer to the ground truth when comparing other models we trained.

\subsection{Framework Structures Experiments}
In this section, based on the results from the first ablation study, we proposed five framework variants, as described in Table \ref{tab:framework}. In order to get a better understanding of the proposed framework in detail from the elimination of the key structure of the full gesture generator.

\begin{table}[!htbp]
\caption{The five framework variants}
\label{tab:framework}
\begin{center}
\begin{tabular}{ l | l }
\toprule
\textbf{Framework} & \textbf{Description}\\
\midrule
Full model & The proposed model \\
No Text & Only used Speech Audio as input \\
No Audio & Only used Speech Text as input  \\
No GRU & Bi-directional GRU layers are not used \\
No FilM Conditions & Previous gesture conditions are not used \\
\bottomrule
\end{tabular}
\end{center}
\end{table}

\begin{table}[!htbp]
\caption{Objective evaluation of proposed frameworks. For Metrics: Closer to the Ground Truth is better. Acceleration(Acc.).}
\label{tab:framework ablation}
\begin{center}
\begin{tabular}{p{1.8cm} p{1.75cm} p{2.1cm} p{1.3cm}}
\toprule
\textbf{Framework} & \textbf{Acc.}$(cm/s^2)$ & \textbf{Jerk}$(cm/s^3)$ & \textbf{RMSE}$(cm)$ \\
\midrule
 Full model & $94.48 \pm 19.64$ & $2187.76 \pm 611.97$ & $4.21 \pm 4.54$\\
\hline
No Text & $105.45 \pm 22.98$ & $2927.35 \pm 686.52$ & $4.23 \pm 4.74$\\
No Audio & $63.01 \pm 10.40$ & $979.33 \pm 162.21$ & $4.46 \pm 4.99$\\
No GRU & $100.36 \pm 22.00$ & $2415.85 \pm 588.61$ & $4.25 \pm 4.78$\\
No FiLM Conditions & $173.44 \pm 37.95$ & $5327.06 \pm 1239.75$ & $4.70 \pm 4.32$\\
\hline
Ground Truth & $144.7 \pm 36.6$ & $2322 \pm 538$ & $0$\\
\bottomrule
\end{tabular}
\end{center}
\end{table}

The results were presented in Table \ref{tab:framework ablation}. Changing any structure of our proposed framework will cause lower results on the RMSE metric.  We note that the results are similar for no GRU compared to the full model, and the reasons could be: 1) The full model may have been too complex for the task. Removing the GRU layer may have resulted in a simpler model that still captures the relevant information from the data. 2) The efficacy of the model may not be significantly affected if the other layers are very good at catching the necessary patterns of the data. In this instance, the lack of the GRU layer might not affect the other layers' ability to accurately reflect the data. Nevertheless, although no-GRU obtained similar results, the full model produced the best overall performance, especially in RMSE. Removing the speech audio input caused higher Jerk than the ground truth while removing the speech text input resulted in the lowest Acceleration among all frameworks.

\section{Conclusion}
We propose a new framework that can generate sequences of joint angles from the speech text and speech audio utterances. Based on a conditional GAN network, the proposed neural network model learns the relationship between the co-speech gestures and both semantic and acoustic features from the speech input. In order to train our neural network model, we employ co-speech gestures with corresponding speech audio utterances dataset, which is captured from a single male native English speaker. Unlike most previous works, our model has the capability to generate continuous gestures associated with the acoustic and semantics of speech.
The results from both objective and subjective evaluations demonstrate the efficacy of our gesture generation framework for robots and embodied agents.

\section*{ACKNOWLEDGMENT}
Thanks to Commonwealth’s contribution for funding this work through an “Australian Government Research Training Program Scholarship”.

\bibliographystyle{IEEEtran}
\bibliography{references}

\end{document}